%% file: neurips_2025.tex
\documentclass{article}

\usepackage{amsmath}

 \usepackage[dblblindworkshop, final]{neurips_2025}
\workshoptitle{Mechanistic Interpretability}

\usepackage[utf8]{inputenc} 
\usepackage[T1]{fontenc}    
\usepackage{hyperref}       
\usepackage{url}            
\usepackage{booktabs}       
\usepackage{amsfonts}       
\usepackage{nicefrac}       
\usepackage{microtype}      
\usepackage{xcolor}         

\usepackage{todonotes}
\usepackage{soul}
\usepackage{pgfplots}
\usepackage{pgfplotstable}
\usepackage{subcaption}

\usepgfplotslibrary{statistics,groupplots}
\pgfplotsset{compat=1.18} 

\title{Activation Transport Operators}

\author{%
  Andrzej Szablewski\thanks{Equal contribution} \\
  University of Cambridge\\
  \texttt{as3623@cam.ac.uk} \\
  \And
  Marek Masiak\footnotemark[1] \\
  University of Oxford\\
  \texttt{marek.masiak@dtc.ox.ac.uk} \\
}

\begin{document}

\maketitle

\begin{abstract}

The residual stream mediates communication between transformer decoder layers via linear reads and writes of non-linear computations.
While sparse-dictionary learning-based methods locate features in the residual stream, and activation patching methods discover circuits within the model, the mechanism by which features \textit{flow} through the residual stream remains understudied. 
Understanding this dynamic can better inform jailbreaking protections, enable early detection of model mistakes, and their correction. 
In this work, we propose Activation Transport Operators (ATO), linear maps from upstream to downstream residuals $k$ layers later, evaluated in feature space using downstream SAE decoder projections. 
We empirically demonstrate that these operators can determine whether a feature has been \textit{linearly transported} from a previous layer or \textit{synthesised} from non-linear layer computation.
We develop the notion of \textit{transport efficiency}, for which we provide an upper bound, and use it to estimate the size of the residual stream subspace that corresponds to linear transport. We empirically demonstrate the linear transport, report transport efficiency and the size of the residual stream's subspace involved in linear transport. 
This compute-light (no finetuning, $<\!50$~GPU-h) method offers practical tools for safety, debugging, and a clearer picture of where computation in LLMs behaves linearly. Our code is available at \url{https://github.com/marek357/activation-transport-operators}.
  
\end{abstract}

\section{Introduction}

Transformer layers modify token-wise residual stream states through a sequence of attention and MLP updates \cite{elhage2021mathematical}. Much of what can be read from these vectors is linear—decoders, probes, and logit-lens all apply affine maps—yet what gets written into the stream is the result of nonlinear mechanisms (LayerNorm, softmax attention, gating in MLPs) \cite{razzhigaev2024transformersecretlylinear}. Many interpretability tools focus either on locating where a behaviour ``lives'' or decoding what a representation ``means'' but they rarely study explicit operators that predict and reconstruct how specific features move from one site in the network to another.

On the intervention side, variants of activation and path patching reliably identify layers, heads, and positions that are causally important for a behaviour \cite{goldowskydill2023localizingmodelbehaviorpath, kramár2024atpefficientscalablemethod}. \citet{ferrando2024informationflowroutesautomatically} present Information Flow Routes, which push further by constructing global, causally validated flow graphs for predictions, yet—like patching—it characterizes influential paths without yielding an explicit map that predicts downstream hidden states. On the decoding side, logit and tuned lenses \cite{nostalgebraist2020logitlens, belrose2025elicitinglatentpredictionstransformers}, provide affine readouts from intermediate residuals into vocabulary space, and sparse autoencoders (SAEs) recover monosemantic features at scale \cite{cunningham2023sparseautoencodershighlyinterpretable}. Furthermore, in their recent study, \citet{lawson2025residualstreamanalysismultilayer} use multi-layer SAEs to study layer similarity, suggesting some evidence of a split between feature transport and non-linear feature recomputation. Meanwhile, activation steering methods demonstrate powerful control via learned activation edits but focus on exogenous behaviour shaping rather than explaining endogenous feature flow \cite{rodriguez2024controllinglanguagediffusionmodels}.

\begin{figure}[t]
    \centering
    \includegraphics[width=0.8\linewidth]{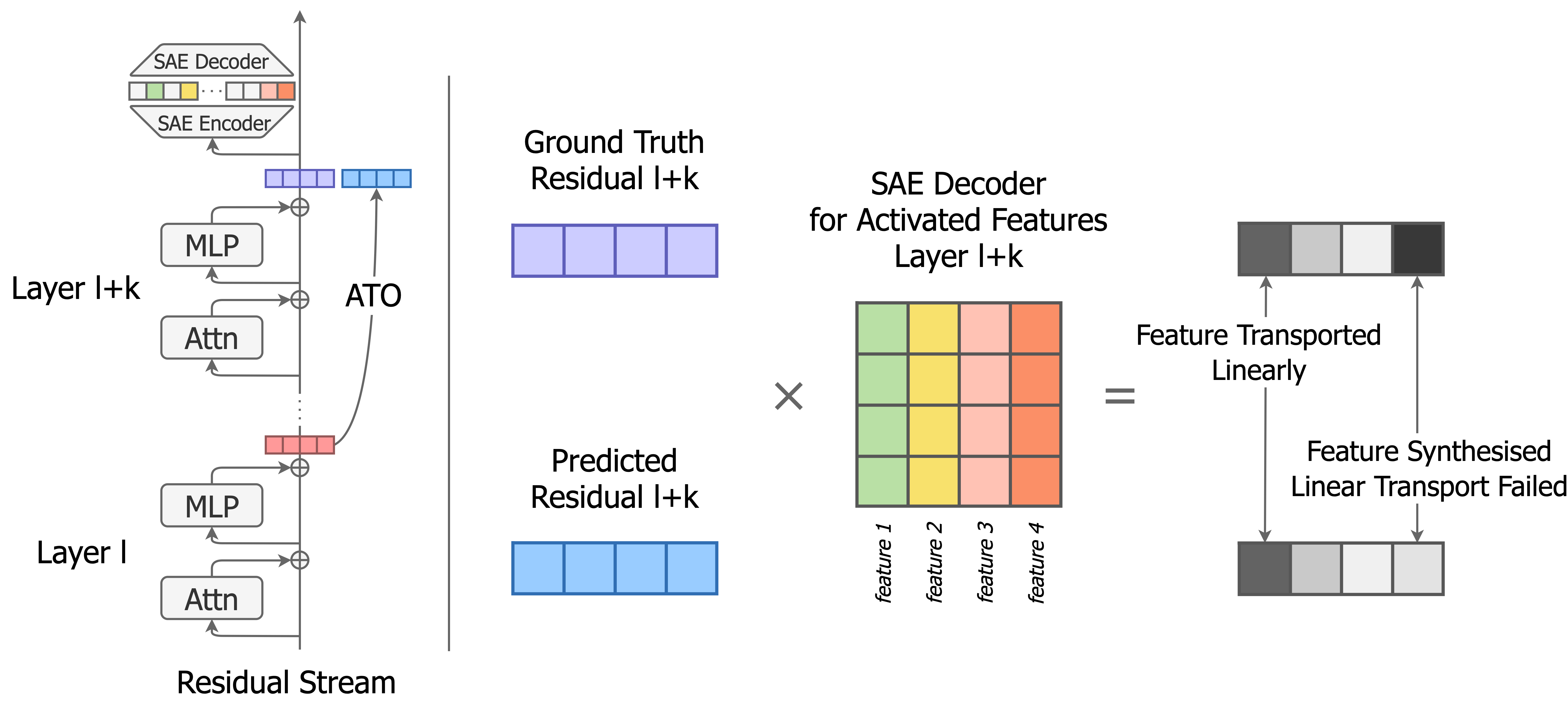}
    \caption{ATO predicts downstream residual stream vector. Using an SAE, we identify activated features. True and predicted residuals are projected onto SAE decoder vectors and compared.}
    \label{fig:diagram}
\end{figure}

This work aims to bridge attribution and representation analysis by introducing Activation Transport Operators (ATOs)—explicit, regularised linear maps that predict downstream residual vectors from upstream residuals. ATOs are learned from paired activations collected during ordinary forward passes. Crucially, ATOs are not a claim that the network is globally linear, but they serve as a test for local linear preservation of a specific feature between two sites in the stream (Figure~\ref{fig:diagram}). High predictive and causal scores indicate linear transport, while failure indicates downstream feature synthesis or nonlinear recomputation.

\textbf{Our core contributions are as follows:} 1) we formally define Activation Transport Operators and empirically study our method using available LLMs and SAEs, evaluating it with per-feature predictive fidelity and causal ablation, and 2) we introduce the notion of transport efficiency, and show its link to the size of the communication subspace of the residual stream.

\section{Methodology}

We study \textit{downstream features} in a decoder-only transformer through the lens of the residual stream. Let $v_{l,i}\in\mathbb{R}^d$ denote the \textit{upstream} residual vector at layer $l$ and token position $i$. For a feature $f$ identified at layer $l{+}k$ by its downstream SAE decoder direction $d_f^{(l+k)}\in\mathbb{R}^d$, the feature is ``observed'' at $(l{+}k, j)$. Our objective is to test whether the downstream activation aligned with $f$ can be \textit{linearly attributed} to earlier residual states. To this end, we learn an affine, rank-constrained transport operator:
\vspace{-0.3cm}

$$
T_r:\mathbb{R}^{d_{\text{model}}}\!\to\!\mathbb{R}^{d_{\text{model}}},\qquad 
\hat v_{l+k,j} \;=\; T_r\,v_{l,i} + b,
$$

where we rank-constrain the transport operator by computing the singular value decomposition:  $T_r = U_r S_r V_r^\top$ with rank $r \leq {d_{\text{model}}}$ (and $b\in\mathbb{R}^{d_{\text{model}}}$). Location pairs $(l,i)\!\to\!(l{+}k,j)$ are sampled using explicit policies, which we refer to as \textit{$j$-policies}. In this work, we use a single $j$-policy: \textit{same-token} ($j{=}i$), which maps upstream to downstream for the same position in a sequence. However, in future work, we plan to explore more complex policies, such as \textit{attention-reader} Top-$K$, \textit{delimiter-pair}, and \textit{copy-target}. The operator is fitted on many such pairs with ridge, lasso, or elasticnet regularisation.
Importantly, evaluation is done in \textit{feature space} rather than on raw residuals. We compare the downstream decoder projections:

\begin{equation}
\label{eq:transport-operator-preds}
a_{\text{true}} \;=\; (d_f^{(l+k)})^\top v_{l+k,j},
\qquad
a_{\text{pred}} \;=\; (d_f^{(l+k)})^\top \hat v_{l+k,j}
\;=\; (d_f^{(l+k)})^\top\!\left(T_r\,v_{l,i}+b\right)
\end{equation}

using regression metrics (specifically, $R^2$ and MSE). High agreement indicates that the component of the downstream state relevant to $f$ is \textit{transported} through a low-dimensional linear channel. On the other hand, poor agreement (despite reasonable upstream sources and policies) suggests the activation is \textit{synthesised locally} by later non-linear computations.

We causally validate the transport operators by ablating the upstream site $(l,i)$ (i.e., zeroing or projecting out the upstream gate) and injecting the reconstructed vector $\hat v_{l+k,j}$ at the target. Restoration of the feature projection and associated behaviour (e.g., structured-format correctness or continuation accuracy) provides direct evidence of linear transport along the learned operator. Additionally, we compare the results with the \textit{zero intervention}, which involves completely ablating the downstream residual vector by setting it to zero \citep{zerointervention1,zerointervention2}. We include this comparison to quantify the maximum corruption we can introduce to the residual stream, thereby measuring the model error (e.g. perplexity increase) if the residual stream contains no information at layer $l{+}k$. We expect this to be significantly larger than the error induced by transport operators.

\paragraph{Transport efficiency}

To better understand the process of feature transport, we seek to find the upper bound for $R^2$ of our rank-$r$ transport operator. Hence, we define the $R^2_{\text{ceiling}}$ as the maximal $R^2$ value achievable by any linear predictor at rank $r$. In this analysis, we shift our focus to the task of predicting downstream residual stream vectors, stacked in matrix $Y \in \mathbb{R}^{N \times d_{\text{model}}}$ from upstream residual stream vectors, stacked in matrix $X \in \mathbb{R}^{N \times d_{\text{model}}}$. Assuming zero-mean, the ceiling for transport efficiency at rank $r$ is given by: $R^2_{\text{ceiling}}(r,\,Y) \;=\; \frac{1}{d_{\text{model}}}\sum_{i=1}^{r} \rho_i^2$, where $\rho_i^2$ are the squared canonical correlations. In Appendix~\ref{ap:r2-ceiling} we rigorously derive this upper bound. 
Therefore, we can define the transport efficiency as:
$\text{Eff} = \tilde{R^2}(r, \hat{Y}_T)/R^2_{\text{ceiling}}(r, Y) \in \left[0,1\right]$,
where $\tilde{R^2}(r, \hat{Y}_T)$ is the $R^2$ metric of rank-$r$-ATO-predicted downstream residual vectors \textit{in whitened Y space}. We need to transform the ATO predictions to the whitened Y space to allow for apples--to--apples comparison of explained variance. Transport efficiency plateaus when increasing ATO's rank does not enhance the relative predictive ability of the operator. This can be observed in Figure \ref{fig:transport} with $k=10$.

\paragraph{Estimating the dimensionality of Linear Transport Subspace (LTS)}
We use the notion of effective dimensionality \citep{effectivedimensionality} to define the dimensionality of the subspace of the residual stream with linear transport: $\mathrm{d}_{\text{eff}} = (\sum_i \rho_i^2)^2/\sum_i (\rho_i^2)^2$.

\paragraph{Experimental setup} We discuss the setup and experimental details in Appendix \ref{ap:experimental-setup}.

\section{Results}

\begin{figure}[h]
    \centering
    \input{./tikz/histograms}
    \caption{Per-feature $R^2$ of operators depend on both the target layer depth and the leap size $k$.}
    \label{fig:histograms}
\end{figure}
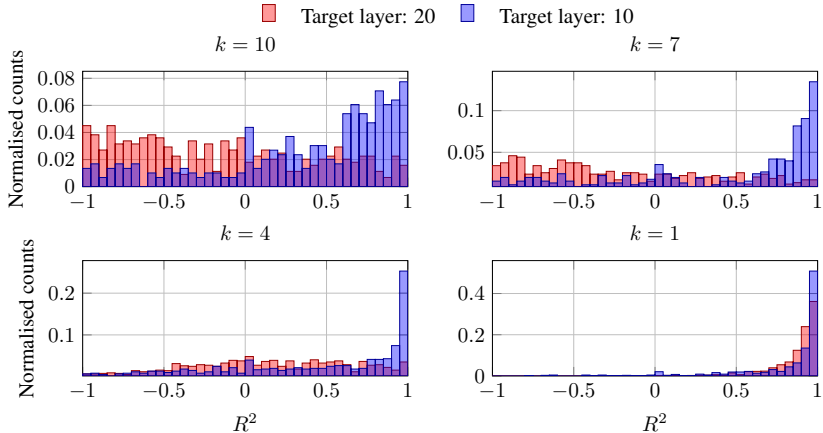

\paragraph{Most linear transport occurs in nearby layers and deteriorates over large distances} 
Comparing the per-feature $R^2$ between full-rank operators shows that those trained for small leaps ($k=1$, $k=4$ for target layer 10) successfully transport a significant number of features ($R^2 > 0.95$). While this number is deteriorating with the growing leap size $k$, we also find that feature transport is generally less common in the later layers of the transformer, even with small $k$s (shown as per-plot distribution shifts in Figure~\ref{fig:histograms}). This suggests that information management in the residual stream may have two regimes. In early layers, the stream has the capacity to accept new features without the need to evict existing ones, hence we observe more transport. Once the residual stream fills up with information, later layers in the model prioritise newly synthesised or non-linearly transformed features, deleting old information from the stream, further supporting the idea introduced by \citet{elhage2021mathematical}.

\begin{figure}[h]
    \centering
    \input{./tikz/heatmap}
    \caption{Average per-feature $R^2$ for all source-target combinations. Note that the sets of chosen SAE features are different across target layers, hence values in the same column may not be directly comparable. Constant leap sizes $k$ are represented by the diagonals.}
    \label{fig:heatmap}
\end{figure}
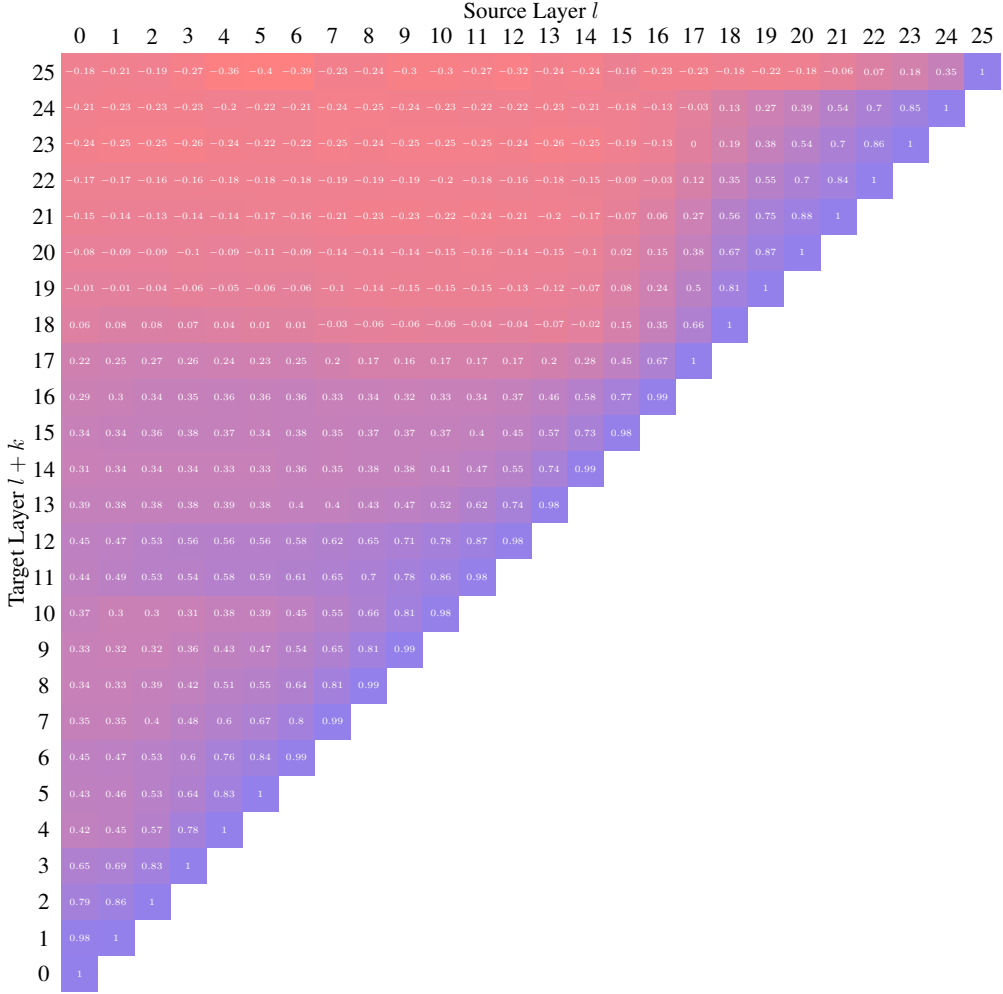

However, we also observe an inverse trend with significantly larger leaps in deeper layers. For example, in layer 21 in Figure \ref{fig:heatmap}, the transport reaches its minimum at $k=10$. Counterintuitively, as the distance between the source and target layer further increases, the $R^2$ metric improves. We find this phenomenon intriguing and will analyse it in detail in further work.

\paragraph{Transport efficiency and LTS size depend on the transport distance} Figure~\ref{fig:transport} shows that transport efficiency over longer leaps ($k{=}7,10$) saturates early and at lower values, indicating a smaller linear transport subspace (i.e. $\text{d}_{\text{eff}}=1453$, and $\text{d}_{\text{eff}}=1291$, respectively). On the other hand, in the adjacent-layer case ($k{=}1$), we observe almost linear improvement of transport efficiency with ATO rank, approaching $R^2_{\text{ceiling}}$ near full rank. Such result is consistent with a larger set of linearly transported directions, size of which is estimated at $\text{d}_{\text{eff}}=2198$. The dimensionality of the LTS should guide ATO rank selection: choosing $r$ above the LTS size yields no population gain beyond the CCA ceiling: extra rank mainly fits noise, which may inflate training $R^2$ but will not generalise.

\paragraph{Using ATOs yields only marginal perplexity increase} We compare perplexity for the unedited, ATO-patched and zero-intervened models. ATOs raise perplexity only slightly, with the effect growing with leap size $k$. The zero-intervened model is significantly worse (similar to using ATO with a null vector), and provides an upper bound on degradation. However, even at $k{=}10$ the increase is 7.1\% of max degradation, and for $k{<}5$, it stays below 1.2\%.  Trends in Figure~\ref{fig:causal} hold beyond the ablations of 5 out of 256 sequence positions; applying ATOs to all positions yields at most a 13.5\% increase at k=10 (with upper-bound perplexity of 12.4529). Thus, ATOs substantially recover language-modelling ability otherwise lost under zero-intervention, supporting their use for targeted diagnostics and edits.

\begin{figure}[h]
\centering
\begin{minipage}[t]{.48\textwidth}
  \vspace{0pt}
  \centering
  \input{./tikz/transport_efficiency_plot}
  \captionof{figure}{Transport efficiency for the target layer~10 with different leap ($k$) values.}
  \label{fig:transport}
\end{minipage}%
\hfill
\begin{minipage}[t]{.48\textwidth}
  \vspace{10pt}
  \centering
  \input{./tikz/causal_plot}
  \captionof{figure}{Log-perplexity for unedited and ablated models. Ablated five positions per sequence.}
  \label{fig:causal}
\end{minipage}
\end{figure}

\paragraph{Limitations} Our study has several limitations. First, we used a single, trivial same-token j-policy, which biases results toward local transport and may miss attention-mediated cross-token routing—exploring IFR-guided or data-driven j selection is left for future work. Second, we evaluated only a single model, therefore, we cannot claim that linear transport is pervasive across architectures or depths without broader replication. Third, our linear operators do not distinguish between features that are transported from earlier layers and those that arise as their linear combinations. Hence, we underestimate the number of synthesised features. Finally, in this work, we do not present feature-targeted editing built with our operators, which we aim to tackle in a follow-up work. In principle, leveraging feature-specific transport between layers could allow low-compute inference-time corrections of the generated text.

\section{Conclusions}

We introduced \emph{Activation Transport Operators} (ATOs): explicit, regularised linear maps that predict a downstream residual vector from upstream residuals and are evaluated in SAE feature space. High predictive and causal scores indicate linear transport of a feature, while failure suggests downstream synthesis or nonlinear recomputation. 
Empirically, we find that transport is strongest over short layer distances and weakens with depth and leap size, suggesting an early-layer regime where the residual stream behaves as a shared linear channel followed by later layers that prioritise synthesis and recomposition. Our transport efficiency metric quantifies how close an operator gets to the best possible linear prediction, while the efficiency analysis implies that the dimensionality of the Linear Transport Subspace is tightly linked to the optimal rank of ATO. 
Taken together, ATOs provide a simple, testable method for mapping feature flow. We expect richer $j$-policies and multi-source operators to reveal attention-mediated routing and to enable feature-targeted, low-compute edits during inference.

\bibliographystyle{plainnat}
\bibliography{bibliography}

\newpage
\appendix

\section{Transport efficiency}
\label{ap:r2-ceiling}

Assuming zero-mean, we define the following covariance matrices:

$$
  \Sigma_{XX}=\tfrac1N X^\top X,\quad
  \Sigma_{YY}=\tfrac1N Y^\top Y,\quad
  \Sigma_{YX}=\tfrac1N Y^\top X,\quad
  \Sigma_{XY}=\Sigma_{YX}^\top.
$$

We employ canonical cross-correlation analysis (CCA) to find directions $a\in\mathbb{R}^{d_{\text{model}}}$ (in downstream residual stream) and $b\in\mathbb{R}^{d_{\text{model}}}$ (in upstream residual stream) maximizing the correlation between the scalar canonical variates, $u = Y a,$ and $v = X b$, subject to $\mathrm{Var}(u)=\mathrm{Var}(v)=1$. Hence, we use the whitening trick to meet the unit variance condition: $\tilde Y \;=\; Y\,\Sigma_{YY}^{-1/2},$ and $
\tilde X \;=\; X\,\Sigma_{XX}^{-1/2}$. Now the covariances of the modified matrices are identities: $\tfrac{1}{N}\tilde Y^\top \tilde Y = I_{d_{\text{model}}}$, $\tfrac{1}{N}\tilde X^\top \tilde X = I_{d_{\text{model}}}$. The whitened cross-covariance is given by $C \;=\; \tfrac{1}{N}\tilde Y^\top \tilde X\;=\;\Sigma_{YY}^{-1/2}\,\Sigma_{YX}\,\Sigma_{XX}^{-1/2}\,\in\mathbb{R}^{d_{\text{model}}\times d_{\text{model}}}$. 

Let the singular value decomposition breakdown of the whitened cross-covariance matrix be $C = U\,\mathrm{diag}(\rho_1,\rho_2,\ldots)\,V^\top$, with singular values $\rho_1\ge \rho_2\ge\cdots\ge 0$. By definition, these $\rho_i$ are the canonical correlations. In other words, in this normalised space, CCA decomposes the relationship between $X$ and $Y$ into orthogonal channels, with each channel strength $\rho_i$, which quantifies how well that specific $Y$ direction can be predicted from $X$. For completeness, the corresponding canonical directions are $a_i \;=\; \Sigma_{YY}^{-1/2} U_{:i}$ and $b_i \;=\; \Sigma_{XX}^{-1/2} V_{:i}$. 

Furthermore, we analyse the matrix $K = CC^\top$. This matrix has the following singular value decomposition: $K = U\,\mathrm{diag}(\rho_i)\,V^\top V\,\mathrm{diag}(\rho_i)\,U^\top
    = U\,\mathrm{diag}(\rho_i^2)\,U^\top$. Importantly, whitening $Y$, implies the optimal linear predictor with rank constraint $r$ captures at most the top-$r$ canonical modes. Therefore, the fraction of explained variance is: $R^2_{\text{ceiling}}(r,\,Y) \;=\; \frac{1}{d_{\text{model}}}\sum_{i=1}^{r} \rho_i^2$.

\section{Experimental setup} \label{ap:experimental-setup}

We conduct experiments using Gemma 2 2B model with hidden dimension $d_{\text{model}} = 2304$, and a suite of pre-trained sparse autoencoders Gemma Scope \cite{gemmateam2024gemma2improvingopen,lieberum2024gemmascopeopensparse}. We use SAEs trained on the post-layer residual stream with the canonical L0 sparsity target and 16,384-dimensional latent space. For training and evaluation of the transport operators, we collect post-layer residual stream hidden states computed over 250,000 tokens from the uniformly subsampled SlimPajama dataset \cite{cerebras2023slimpajama}, available under Apache 2.0 license. We subsequently split the dataset into 60\% train, 20\% validation and 20\% test splits. For each layer, we identify $\sim\!5\%$ high-quality SAE features, which we use in the operator evaluation by processing 120,000 dataset tokens and applying heuristics preferring features with high semantic coherence (low token entropy), centred probability mass in the unembedding space projections, as well as most significant causal effects. Furthermore, we filter out highly redundant and dead features. 

To study the dynamics of feature transport throughout the model, we investigate target decoder layers $10$ and $20$ and compare the reconstruction of the same set of features per target layer, offset by $k=\{1, \dots, 9\}$. Additionally, we ablate over all target layers andthe leap size to create the heatmap shown in Figure \ref{fig:heatmap}. We implement transport operators as $L_2$-regularised ridge regression models, trained using 5-fold cross-validation with grid search over regularisation parameter $\alpha$, and choose a model with the highest $R^2$ score. To evaluate the models, we measure the reconstructions of transport operators with regards to the selected SAE features. To address the inherent sparsity of SAE features, we ensure predicting only activated latents. Furthermore, we analyse only those, which activated at least ten times in the test dataset and achieved $R^2 > -1$. 

In the transport efficiency study, we evaluate transport operators by computing whitened $R^2$ of the rank-$r$-ATO-predicted downstream residuals, for all values $r$ starting with 1 and incremented by 50 until $d_{\text{model}}$.

In the causal validation, we compare the unedited and ablated models by computing perplexity over a held-out subset over 100 sequences of 256 tokens. We experiment with 3 configurations of distinct token positions, to which the modification is applied: only one position, five positions, and all positions in a sequence. In the first two cases, we randomly choose positions from throughout the sequence and average the resulting perplexity over 3 sets of positions for robustness. We perform all computation in single precision (\verb|float32|) using M1 Pro and M2 Max hardware. 

\end{document}

%% file: tikz/histograms.tex
\begin{tikzpicture}[scale=0.8, transform shape]
\begin{groupplot}[
  group style={
    group size=2 by 2,
    horizontal sep=40pt,
    vertical sep=35pt,
    ylabels at=edge left,
    xlabels at=edge bottom                        
  },
  width=0.5\textwidth,
  height=3.5cm,
  xlabel={$R^2$},
  ylabel={Normalised counts},
  ymajorgrids,
  xmajorgrids,
  enlarge x limits=0,
  enlarge y limits={rel=0.1,upper},
  legend style={draw=none, fill=none, column sep=1em},
  legend columns=-1,
  every axis plot/.append style={ybar legend},  
  scaled y ticks=false,
  yticklabel style={/pgf/number format/fixed, /pgf/number format/precision=4},
]

\nextgroupplot[title={$k=10$},
  legend to name=grouplegend                  
]
\pgfplotstablegetrowsof{./plots_data/histograms/r2_layer_10_k_10_r2_lat.csv}
\pgfmathsetmacro{\N}{\pgfplotsretval} 
\addplot+[
  ybar interval,
  mark=none,
  hist={bins=40, data min=-1, data max=1},
  fill=red, draw=red!60!black,
  fill opacity=0.40,
  y filter/.expression={y/\N} 
] table [col sep=comma, y=value] {./plots_data/histograms/r2_layer_10_k_10_r2_lat.csv};
\addlegendentry{Target layer: 20}
\pgfplotstablegetrowsof{./plots_data/histograms/r2_layer_0_k_10_r2_lat.csv}
\pgfmathsetmacro{\N}{\pgfplotsretval} 
\addplot+[
  ybar interval,
  mark=none,
  hist={bins=40, data min=-1, data max=1},
  fill=blue, draw=blue!60!black,
  fill opacity=0.40,
  y filter/.expression={y/\N} 
] table [col sep=comma, y=value] {./plots_data/histograms/r2_layer_0_k_10_r2_lat.csv};
\addlegendentry{Target layer: 10}

\nextgroupplot[title={$k=7$}]
\pgfplotstablegetrowsof{./plots_data/histograms/r2_layer_13_k_7_r2_lat.csv}
\pgfmathsetmacro{\N}{\pgfplotsretval} 
\addplot+[
  ybar interval,
  mark=none,
  hist={bins=40, data min=-1, data max=1},
  fill=red, draw=red!60!black,
  fill opacity=0.40,
    y filter/.expression={y/\N} 
] table [col sep=comma, y=value] {./plots_data/histograms/r2_layer_13_k_7_r2_lat.csv};
\pgfplotstablegetrowsof{./plots_data/histograms/r2_layer_3_k_7_r2_lat.csv}
\pgfmathsetmacro{\N}{\pgfplotsretval} 
\addplot+[
  ybar interval,
  mark=none,
  hist={bins=40, data min=-1, data max=1},
  fill=blue, draw=blue!60!black,
  fill opacity=0.40,
  y filter/.expression={y/\N} 
] table [col sep=comma, y=value] {./plots_data/histograms/r2_layer_3_k_7_r2_lat.csv};

\nextgroupplot[title={$k=4$}]
\pgfplotstablegetrowsof{./plots_data/histograms/r2_layer_16_k_4_r2_lat.csv}
\pgfmathsetmacro{\N}{\pgfplotsretval} 
\addplot+[
  ybar interval,
  mark=none,
  hist={bins=40, data min=-1, data max=1},
  fill=red, draw=red!60!black,
  fill opacity=0.40,
     y filter/.expression={y/\N} 
] table [col sep=comma, y=value] {./plots_data/histograms/r2_layer_16_k_4_r2_lat.csv};
\pgfplotstablegetrowsof{./plots_data/histograms/r2_layer_6_k_4_r2_lat.csv}
\pgfmathsetmacro{\N}{\pgfplotsretval} 
\addplot+[
  ybar interval,
  mark=none,
  hist={bins=40, data min=-1, data max=1},
  fill=blue, draw=blue!60!black,
  fill opacity=0.40,
    y filter/.expression={y/\N} 
] table [col sep=comma, y=value] {./plots_data/histograms/r2_layer_6_k_4_r2_lat.csv};

\nextgroupplot[title={$k=1$}]
\pgfplotstablegetrowsof{./plots_data/histograms/r2_layer_19_k_1_r2_lat.csv}
\pgfmathsetmacro{\N}{\pgfplotsretval} 
\addplot+[
  ybar interval,
  mark=none,
  hist={bins=40, data min=-1, data max=1},
  fill=red, draw=red!60!black,
  fill opacity=0.40,
    y filter/.expression={y/\N} 
] table [col sep=comma, y=value] {./plots_data/histograms/r2_layer_19_k_1_r2_lat.csv};
\pgfplotstablegetrowsof{./plots_data/histograms/r2_layer_9_k_1_r2_lat.csv}
\pgfmathsetmacro{\N}{\pgfplotsretval} 
\addplot+[
  ybar interval,
  mark=none,
  hist={bins=40, data min=-1, data max=1},
  fill=blue, draw=blue!60!black,
  fill opacity=0.40,
    y filter/.expression={y/\N} 
] table [col sep=comma, y=value] {./plots_data/histograms/r2_layer_9_k_1_r2_lat.csv};

\end{groupplot}
\node at ($(group c1r1.north)!0.5!(group c2r1.north)$) [yshift=0.9cm]
  {\pgfplotslegendfromname{grouplegend}};
  
\end{tikzpicture}

%% file: tikz/heatmap.tex
\newcommand{\cellsize}{8mm}        
\newcommand{\xstep}{8mm}           
\newcommand{\ystep}{8mm}           
\newcommand{\labelfont}{\Large}     
\newcommand{\cellfont}{\tiny}

\definecolor{mapred}{RGB}{215,48,39}    
\definecolor{mapgreen}{RGB}{35,132,67}  

\scalebox{0.6}{%

\begin{tikzpicture}[scale=1, x=\xstep, y=\ystep]
  \foreach \col in {0,...,25} {
    \node[minimum size=\cellsize, font=\labelfont, inner sep=0pt] at (\col+1, 0) {\col};
  }

  \foreach \row in {0,...,25} {
    \pgfmathsetmacro{\yy}{-(26 - \row)}
    \node[minimum size=\cellsize, font=\labelfont, inner sep=0pt] at (0,\yy) {\row};
  }

  \foreach \x/\y/\c in {
  0/25/1.000000,
  0/24/0.983011, 1/24/1.000000,
  0/23/0.785688, 1/23/0.862972, 2/23/1.000000,
  0/22/0.653980, 1/22/0.687456, 2/22/0.828553, 3/22/1.000000,
  0/21/0.423903, 1/21/0.448439, 2/21/0.573402, 3/21/0.780136, 4/21/0.998586,
  0/20/0.425035, 1/20/0.457292, 2/20/0.531026, 3/20/0.635648, 4/20/0.828885, 5/20/0.995370,
  0/19/0.452574, 1/19/0.466569, 2/19/0.529013, 3/19/0.604625, 4/19/0.756937, 5/19/0.843641, 6/19/0.985849,
  0/18/0.353393, 1/18/0.347876, 2/18/0.399036, 3/18/0.476126, 4/18/0.595795, 5/18/0.670526, 6/18/0.804824, 7/18/0.994269,
  0/17/0.335138, 1/17/0.329737, 2/17/0.388394, 3/17/0.420250, 4/17/0.509074, 5/17/0.547078, 6/17/0.641126, 7/17/0.808162, 8/17/0.992816,
  0/16/0.332939, 1/16/0.316900, 2/16/0.322815, 3/16/0.361752, 4/16/0.432908, 5/16/0.467113, 6/16/0.544886, 7/16/0.646303, 8/16/0.810852, 9/16/0.992898,
  0/15/0.365125, 1/15/0.302793, 2/15/0.300220, 3/15/0.308874, 4/15/0.379131, 5/15/0.387300, 6/15/0.448170, 7/15/0.554224, 8/15/0.664270, 9/15/0.814038, 10/15/0.981050,
  0/14/0.441111, 1/14/0.489467, 2/14/0.525431, 3/14/0.539427, 4/14/0.582235, 5/14/0.585552, 6/14/0.614602, 7/14/0.653619, 8/14/0.703857, 9/14/0.775659, 10/14/0.862297, 11/14/0.980142,
  0/13/0.446308, 1/13/0.465577, 2/13/0.530276, 3/13/0.558194, 4/13/0.564532, 5/13/0.559115, 6/13/0.584946, 7/13/0.616001, 8/13/0.652417, 9/13/0.706904, 10/13/0.779962, 11/13/0.868908, 12/13/0.977931,
  0/12/0.386608, 1/12/0.378399, 2/12/0.379480, 3/12/0.378314, 4/12/0.387189, 5/12/0.378068, 6/12/0.398012, 7/12/0.402802, 8/12/0.430052, 9/12/0.471982, 10/12/0.524324, 11/12/0.619823, 12/12/0.744557, 13/12/0.984680,
  0/11/0.313216, 1/11/0.335834, 2/11/0.340118, 3/11/0.339050, 4/11/0.330760, 5/11/0.334807, 6/11/0.359635, 7/11/0.353743, 8/11/0.377259, 9/11/0.379061, 10/11/0.410023, 11/11/0.468178, 12/11/0.545023, 13/11/0.735910, 14/11/0.988858,
  0/10/0.336059, 1/10/0.343843, 2/10/0.364135, 3/10/0.375530, 4/10/0.373276, 5/10/0.344591, 6/10/0.379702, 7/10/0.354047, 8/10/0.368770, 9/10/0.365807, 10/10/0.373542, 11/10/0.400234, 12/10/0.449852, 13/10/0.569646, 14/10/0.731573, 15/10/0.984021,
  0/9/0.290157, 1/9/0.301293, 2/9/0.344542, 3/9/0.347209, 4/9/0.362152, 5/9/0.355070, 6/9/0.359842, 7/9/0.327580, 8/9/0.335735, 9/9/0.321111, 10/9/0.329888, 11/9/0.338944, 12/9/0.374625, 13/9/0.455888, 14/9/0.581628, 15/9/0.772716, 16/9/0.989796,
  0/8/0.219849, 1/8/0.249405, 2/8/0.267966, 3/8/0.263593, 4/8/0.242960, 5/8/0.232698, 6/8/0.249591, 7/8/0.195495, 8/8/0.165711, 9/8/0.160259, 10/8/0.173988, 11/8/0.174234, 12/8/0.166631, 13/8/0.196811, 14/8/0.275142, 15/8/0.454676, 16/8/0.668040, 17/8/0.998711,
  0/7/0.056552, 1/7/0.081769, 2/7/0.083030, 3/7/0.066237, 4/7/0.044979, 5/7/0.009489, 6/7/0.011677, 7/7/-0.031144, 8/7/-0.056999, 9/7/-0.058753, 10/7/-0.056062, 11/7/-0.043757, 12/7/-0.039244, 13/7/-0.067510, 14/7/-0.018684, 15/7/0.151258, 16/7/0.345966, 17/7/0.655285, 18/7/0.998748,
  0/6/-0.013122, 1/6/-0.010393, 2/6/-0.039367, 3/6/-0.062999, 4/6/-0.045375, 5/6/-0.058974, 6/6/-0.058229, 7/6/-0.102647, 8/6/-0.138908, 9/6/-0.154233, 10/6/-0.153365, 11/6/-0.148109, 12/6/-0.134570, 13/6/-0.119613, 14/6/-0.072404, 15/6/0.078484, 16/6/0.243258, 17/6/0.502882, 18/6/0.809255, 19/6/0.997519,
  0/5/-0.075151, 1/5/-0.089011, 2/5/-0.087882, 3/5/-0.096543, 4/5/-0.089898, 5/5/-0.106542, 6/5/-0.092579, 7/5/-0.139622, 8/5/-0.136850, 9/5/-0.139540, 10/5/-0.149368, 11/5/-0.158921, 12/5/-0.142593, 13/5/-0.152506, 14/5/-0.102865, 15/5/0.016509, 16/5/0.153432, 17/5/0.383256, 18/5/0.667595, 19/5/0.870054, 20/5/1.000000,
  0/4/-0.151826, 1/4/-0.139253, 2/4/-0.126536, 3/4/-0.138604, 4/4/-0.144665, 5/4/-0.170719, 6/4/-0.164239, 7/4/-0.206059, 8/4/-0.225531, 9/4/-0.233225, 10/4/-0.222542, 11/4/-0.235683, 12/4/-0.210846, 13/4/-0.200533, 14/4/-0.167468, 15/4/-0.072117, 16/4/0.063258, 17/4/0.272202, 18/4/0.556981, 19/4/0.748173, 20/4/0.881893, 21/4/0.997540,
  0/3/-0.169143, 1/3/-0.171980, 2/3/-0.164872, 3/3/-0.160078, 4/3/-0.179025, 5/3/-0.179039, 6/3/-0.183870, 7/3/-0.191445, 8/3/-0.188518, 9/3/-0.193844, 10/3/-0.201078, 11/3/-0.184127, 12/3/-0.163851, 13/3/-0.177078, 14/3/-0.149586, 15/3/-0.094455, 16/3/-0.030273, 17/3/0.120168, 18/3/0.352481, 19/3/0.549217, 20/3/0.697071, 21/3/0.840461, 22/3/1.000000,
  0/2/-0.237404, 1/2/-0.246105, 2/2/-0.252079, 3/2/-0.261501, 4/2/-0.240517, 5/2/-0.219049, 6/2/-0.222969, 7/2/-0.251468, 8/2/-0.238651, 9/2/-0.245152, 10/2/-0.245901, 11/2/-0.250689, 12/2/-0.235967, 13/2/-0.264498, 14/2/-0.249193, 15/2/-0.187360, 16/2/-0.128613, 17/2/0.002712, 18/2/0.191557, 19/2/0.380614, 20/2/0.538356, 21/2/0.697426, 22/2/0.860424, 23/2/1.000000,
  0/1/-0.212993, 1/1/-0.229140, 2/1/-0.228666, 3/1/-0.227608, 4/1/-0.197427, 5/1/-0.224967, 6/1/-0.211710, 7/1/-0.243889, 8/1/-0.254061, 9/1/-0.239830, 10/1/-0.225549, 11/1/-0.220772, 12/1/-0.221522, 13/1/-0.232351, 14/1/-0.207460, 15/1/-0.179690, 16/1/-0.125857, 17/1/-0.031381, 18/1/0.126262, 19/1/0.267437, 20/1/0.389250, 21/1/0.542726, 22/1/0.696257, 23/1/0.853162, 24/1/1.000000,
  0/0/-0.182103, 1/0/-0.209417, 2/0/-0.193536, 3/0/-0.271774, 4/0/-0.359851, 5/0/-0.399500, 6/0/-0.386818, 7/0/-0.233228, 8/0/-0.236941, 9/0/-0.303322, 10/0/-0.298329, 11/0/-0.272016, 12/0/-0.317301, 13/0/-0.241659, 14/0/-0.243083, 15/0/-0.157000, 16/0/-0.229574, 17/0/-0.228148, 18/0/-0.179162, 19/0/-0.219854, 20/0/-0.184880, 21/0/-0.059099, 22/0/0.067146, 23/0/0.183385, 24/0/0.349175, 25/0/1.000000}
  {
    \pgfmathsetmacro{\p}{round(60*(\c+0.4))} 
    \pgfmathsetmacro{\yy}{-(26 - \y)}
    \node[
      fill=blue!\p!red!50,
      minimum size=\cellsize,
      text=white,
      font=\cellfont,
      inner sep=0pt
    ] at (\x+1,-\y-1) {\pgfmathprintnumber[fixed,precision=2]{\c}};
  }

  \node[below, font=\labelfont] at (13.5,1.1) {Source Layer $l$};
  \node[rotate=90, font=\labelfont] at (-0.7,-13.5) {Target Layer $l + k$};

\end{tikzpicture}
}

%% file: tikz/transport_efficiency_plot.tex


\begin{tikzpicture}[baseline=(current bounding box.north)]
\begin{axis}[
  width=\linewidth, height=0.6\linewidth,
  xlabel={Operator rank}, ylabel={Transport Efficiency},
  ymajorgrids, xmajorgrids,
  enlarge x limits = false,
  axis y line=left, axis x line=bottom,
  scaled y ticks=false,
  yticklabel style={/pgf/number format/fixed, /pgf/number format/precision=4},
  ymin=0, ymax=1,
  xmin=0,
  legend style={nodes={scale=0.6, transform shape}, at={(0.49,0.92)},anchor=south, legend columns=4},
]

\addplot[thick, mark=none, blue, mark options={scale=0.7}] table[x=r, y=Efficiency, col sep=comma, header=true]{./plots_data/transport_efficiency/L9_k1.csv};
\addlegendentry{$k{=}1$}
\addplot[thick, mark=none, red, mark options={scale=0.5}] table[x=r, y=Efficiency, col sep=comma, header=true]{./plots_data/transport_efficiency/L6_k4.csv};
\addlegendentry{$k{=}3$}
\addplot[thick, mark=none, cyan, mark options={scale=0.7}] table[x=r, y=Efficiency, col sep=comma, header=true]{./plots_data/transport_efficiency/L3_k7.csv};
\addlegendentry{$k{=}7$}
\addplot[thick, mark=none, brown, mark options={scale=0.7}] table[x=r, y=Efficiency, col sep=comma, header=true]{./plots_data/transport_efficiency/L0_k10.csv};
\addlegendentry{$k{=}10$}

\end{axis}
\end{tikzpicture}

%% file: tikz/causal_plot.tex
\usepgfplotslibrary{fillbetween}

\pgfplotstableread[col sep=comma, header=true]{./plots_data/causal/aggregated_perplexity_js5.csv}\perptable

\begin{tikzpicture}[baseline=(current bounding box.north)]
\begin{axis}[
  width=\linewidth, height=0.6\linewidth,
  xlabel={$k$}, ylabel={$\log\mathrm{PPL}$ (↓)},
  ymajorgrids, xmajorgrids,
  enlarge x limits = false,
  axis y line=left, axis x line=bottom,
  scaled y ticks=false,
  yticklabel style={/pgf/number format/fixed, /pgf/number format/precision=4},
  xtick distance = 1,
  legend style={nodes={scale=0.6, transform shape}, at={(0.63,0.25)},anchor=south, legend},
  ymax=2.545,
  ymin=2.447,
]

\addplot[thick, mark=square*, red, mark options={scale=0.9}] table[x=k, y=ablation]{\perptable};
\addlegendentry{ATO from null vector (95\% CI)}
\addplot[thick, mark=*,  blue, mark options={scale=0.9}] table[x=k, y=intervention]{\perptable};
\addlegendentry{ATO from upstream  (95\% CI)}

\addplot[name path=abla_up,   draw=none, forget plot] table[x=k, y=ablation_ci_upper]{\perptable};
\addplot[name path=abla_down, draw=none, forget plot] table[x=k, y=ablation_ci_lower]{\perptable};
\addplot[red!25, opacity=0.6, forget plot] fill between[of=abla_up and abla_down];

zero-intervention (ceiling)
\addplot[very thick, black, dotted, mark=none, mark options={scale=0.9}] table[x=k, y=zero]{\perptable};
\addlegendentry{Zero intervention}

\addplot[name path=interv_up,   draw=none, forget plot] table[x=k, y=intervention_ci_upper]{\perptable};
\addplot[name path=interv_down, draw=none, forget plot] table[x=k, y=intervention_ci_lower]{\perptable};
\addplot[blue!25, opacity=0.6, forget plot] fill between[of=interv_up and interv_down];

\addplot[very thick, black, dashed, mark=none, mark options={scale=0.9}] table[x=k, y=baseline]{\perptable};
\addlegendentry{Unedited model}

\end{axis}
\end{tikzpicture}